\pdfoutput=1

\documentclass[11pt]{article}

\usepackage{EMNLP2023}

\usepackage{times}
\usepackage{latexsym}
\usepackage{amsmath}
\usepackage{graphicx}
\usepackage{pifont}
\usepackage{booktabs}

\usepackage[T1]{fontenc}

\usepackage[utf8]{inputenc}

\usepackage{microtype}

\usepackage{inconsolata}

\title{LLMs are Frequency Pattern Learners in Natural Language Inference}

\author{{\bf Liang Cheng}$^{\dag}$ \quad {\bf Zhaowei Wang}$^{\ddag}$ \quad {\bf Mark Steedman}$^{\dag}$ \\
        $^{\dag}$University of Edinburgh \quad
        $^{\ddag}$HKUST \quad \\
        \texttt{L.Cheng@ed.ac.uk \quad zwanggy@cse.ust.hk} \quad m.steedman@ed.ac.uk
        }

\begin{document}
\maketitle
\begin{abstract}
While fine-tuning LLMs on NLI corpora improves their inferential performance, the underlying mechanisms driving this improvement remain largely opaque.
In this work, we conduct a series of experiments to investigate what LLMs actually learn during fine-tuning. 
We begin by analyzing predicate frequencies in premises and hypotheses across NLI datasets and identify a consistent \textbf{frequency bias}, where predicates in hypotheses occur more frequently than those in premises for positive instances.
To assess the impact of this bias, we evaluate both standard and NLI fine-tuned LLMs on bias-consistent and bias-adversarial cases. We find that LLMs exploit frequency bias for inference and perform poorly on adversarial instances. Furthermore, fine-tuned LLMs exhibit significantly increased reliance on this bias, suggesting that they are learning these frequency patterns from datasets. 
Finally, we compute the frequencies of hyponyms and their corresponding hypernyms from WordNet, revealing a correlation between frequency bias and textual entailment. These findings help explain why learning frequency patterns can enhance model performance on inference tasks.


\end{abstract}

\section{Introduction}
\label{sec:introduction}
Natural Language Inference (NLI) is a core task in language understanding, aiming to determine whether a hypothesis follows logically from a premise.
The rise of LLMs has driven significant progress in NLI \cite{he2024using, liu2024large}, with studies \cite{liu2020adversarial, li2025dynmole} showing that training on NLI corpora improves performance on inference benchmarks. 
Moreover, \citet{cheng2025neutralizing} introduce counterfactual NLI datasets to train LLMs, significantly enhancing their inferential abilities while reducing hallucinations. 

Despite the widespread use of NLI datasets for training, the underlying mechanisms by which LLMs acquire inferential capabilities remain unclear. 
\citet{li2022language} argue that the performance gains of LLMs from inference data result from overfitting to dataset artifacts. \citet{mckenna2023sources} prove that LLMs benefit from memorizing such artifacts and leveraging them as shortcuts for inference.
However, these studies fail to account for the finding of \citet{cheng2025neutralizing} that training on counterfactual reasoning data can still lead to improved inference performance.
To further understand the source of LLMs' performance gains from training on NLI corpora, we conduct a series of controlled experiments analyzing their inferential behavior.

\textit{First}, we examine the frequency of predicates in premises and hypotheses separately across multiple NLI datasets to analyze their underlying distributional properties. Our experiments reveal a clear \textbf{frequency biases} in NLI datasets: predicates in hypotheses tend to occur with significantly higher frequency than those in premises in examples labeled as positive. 
\textit{Second}, we evaluate the performance of both standard and NLI-tuned LLMs on inference benchmarks, presenting evidence that these models are sensitive to frequency bias and tend to rely on this bias during inference. We further prove that fine-tuning on NLI datasets amplifies this reliance on frequency bias. Furthermore, we partition the NLI test set based on whether samples are \textit{consistent} with or \textit{adversarial} to frequency bias. Evaluation results show that models fine-tuned on inference data perform poorly on frequency-adversarial  inference, proving that training on NLI datasets leads models to learn the frequency bias from NLI datasets.
\textit{Finally}, we present experiments demonstrating a correlation between frequency bias and textual entailment, offering an explanation for why learning frequency bias may serve as a proxy for enhancing inferential capability. 

\begin{table*}[!t]
 \vspace{-0.1in}
    \centering\resizebox{0.7\textwidth}{!}{
   \begin{tabular}{ccccccc}
\hline
          & \multicolumn{3}{c}{\textbf{Label =} \texttt{Entail}}                    & \multicolumn{3}{c}{\textbf{Label =} \texttt{No-Entail}}                     \\ \hline
          & premise& hypotheiss& \multicolumn{1}{l|}{\textit{more frequent}} & premise& hypothesis& \multicolumn{1}{l}{\textit{more frequent}} \\ \cline{2-7} 
EGs       & 36.05& 209.1&  \textcolor{red}{hypothesis}& 109.35& 32.84& \textcolor{blue}{premise}\\
Levy/Holt & 9.97& 24.29& \textcolor{red}{hypothesis}& 8.38& 6.53& \textcolor{blue}{premise}\\ 
 RTE& 60.89& 62.24& \textcolor{red}{hypothesis}& 73.01& 60.09&\textcolor{blue}{premise}\\
 MNLI& 69.30& 72.71& \textcolor{red}{hypothesis}& 69.10& 65.80&\textcolor{blue}{premise}\\ \hline
\end{tabular}
}
    \caption{Average predicate frequency in \textcolor{red}{hypothesis} and \textcolor{blue}{premise} across different datasets. Frequencies are computed separately for positive (\texttt{Entail}) and negative (\texttt{No-Entail}) examples. }
    \label{tab:average_frequency_across_P-H}
    \vspace{-0.1in}
\end{table*}
The main contributions of this paper are summarized as follows:

(a) We identify frequency bias in the various NLI data sets, where hypotheses tend to occur more frequently than premises when labeled is positive.

(b) We evaluate a range of LLMs and their fine-tuned counterparts, demonstrating that these models tend to exploit frequency bias during inference, and that fine-tuning on NLI datasets further amplifies this reliance. 

(c) We present the relation between frequency bias and textual entailment, providing explanations why learning frequency bias can improve inference performance.

\section{Methods}
\label{sec:methods}
\subsection{Calculate average frequency of predicates}
\label{sec:calculate frequency}
For each hypothesis and premise in NLI datsets, we compute n-gram frequencies using the \textit{WordFreq} library \cite{robyn_speer_2022_7199437}, which aggregates frequency data from multiple text sources to provide reliable usage statistics. In our experiments, we focus exclusively on the frequency of \textit{verbal} predicates, abstracting away from the specific entity arguments that accompany them. 
We calculate the average predicate frequency for each statement. This approach enables us to capture general usage patterns of predicates across corpora while minimizing effects introduced by different subjects or objects\footnote{https://github.com/LeonChengg/FrequencyBias-Analysis}.

\subsection{Metrics}
\label{sec:metrics}
We calculate frequency bias as the difference between the frequencies of the hypothesis and the premise, computed as follows:
\begin{align*}
 \vspace{-0.2in}
    \textit{Bias}(hypo, prem) = \textit{Freq}(hypo) - \textit{Freq}(prem)
 \vspace{-0.2in}
\end{align*}
where \textit{Freq}($\cdot$) denotes the average frequency of predicates in a given statement\footnote{To better illustrate the frequency differences, we scale all frequency values by a factor of 1,000. }.

\section{Experimental Setup}
\label{sec:experiments}
\subsection{Datasets}
\label{sec:datasets}
We fine-tune LLMs on a range of widely used NLI datasets and evaluate their inferential performance. The training sets include \textbf{RTE} \cite{dagan2006pascal, wang2019glue}, \textbf{MNLI} \cite{williams2018broad, wang2019glue}, and \textbf{Entailment Graphs (EGs)} \cite{hosseini2018learning,hosseini2021open, cheng2025neutralizing}, which is a counterfactual yet logically valid reasoning dataset. For evaluation, we use the \textbf{Levy/Holt} \cite{levy_annotating_2016, holt_probabilistic_2019} dataset, where each premise–hypothesis pair contains a single predicate with two named entity arguments, allowing for clear analysis of predicate frequency patterns.

We adopt the same prompt templates used in prior work \cite{schmitt_language_2021, mckenna2023sources, cheng2025neutralizing} for both fine-tuning and inference, which format samples as binary questions to determine whether the premise entails the hypothesis. A positive label corresponds to \texttt{Entail}, and a negative label to \texttt{No-Entail}.
Details of the datasets and prompt configurations are provided in Appendix~\ref{sec:appendix_dataset} and \ref{sec:appendix_prompts}. 

\begin{table*}[h]
    \centering\resizebox{1\textwidth}{!}{
   \begin{tabular}{ccccccccc}
\hline
           &  &\multicolumn{3}{c}{\textbf{Label =} \texttt{Entail}}             &  &\multicolumn{3}{c}{\textbf{Label = }\texttt{No-Entail}}            \\ \hline
           & premise& hypothesis&  \textit{Bias(\textcolor{red}{hypo},\textcolor{blue}{prem})}&\multicolumn{1}{l|}{\textit{follow bias}} & premise& hypothesis&  \textit{Bias(\textcolor{blue}{prem},\textcolor{red}{hypo})}&\multicolumn{1}{l}{\textit{follow bias}} \\ 
           \cline{2-9} DeepSeek-8B \textcolor{green}{\ding{51}}&   20.27&  65.84&  45.57&consistent&  88.98&  11.75&  77.23&consistent\\
 DeepSeek-8B \textcolor{red}{\ding{55}}& 8.27  & 52.59 &  \textbf{44.32}&consistent& 40.83 & 24.15 & \textbf{16.68}&consistent\\ 
 \cline{2-9}LLaMA-3-8B \textcolor{green}{\ding{51}}& 17.58&  66.18&  48.6&consistent& 112.15& 13.36& 98.79&consistent\\
 LLaMA-3-8B \textcolor{red}{\ding{55}}& 26.28& 33.63&  \textbf{7.35}&consistent& 45.95& 20.07& \textbf{25.88}&consistent\\
 \cline{2-9}Mistral-7B \textcolor{green}{\ding{51}}& 20.06& 66.83&  46.77&consistent& 92.87& 18.31& 74.56&consistent\\
 Mistral-7B \textcolor{red}{\ding{55}}& 11.02& 50.71&  \textbf{39.69}&consistent& 36.39& 18.25& \textbf{18.14}&consistent\\
 \cline{2-9}LLaMA-3-70B \textcolor{green}{\ding{51}}& 17.13& 67.96&  50.83&consistent& 77.56& 19.37& 58.19&consistent\\
 LLaMA-3-70B \textcolor{red}{\ding{55}}& 24.66& 39.15&  \textbf{14.49}&consistent& 44.03& 16.75& \textbf{27.28}&consistent\\ \hline
 DeepSeek-8B$_{EG}$ \textcolor{green}{\ding{51}}& 15.78& 79.85&  64.07&consistent& 95.18& 8.26& 86.92&consistent\\
DeepSeek-8B$_{EG}$ \textcolor{red}{\ding{55}}& 25.64 & 15.55 &  \textbf{-10.09}&\textit{\underline{adversarial}}& 26.50 & 30.06 &  \textbf{-3.56}&\textit{\underline{adversarial}}\\ 
 \cline{2-9}LLaMA-3-8B$_{EG}$ \textcolor{green}{\ding{51}}& 14.92& 66.27&  51.35&consistent& 130.86& 8.26& 122.6&consistent\\
 LLaMA-3-8B$_{EG}$ \textcolor{red}{\ding{55}}&  69.18& 23.45&  \textbf{-45.73}&\textit{\underline{adversarial}}& 38.01& 22.10& \textbf{15.91}&consistent\\
 \cline{2-9}Mistral-7B$_{EG}$ \textcolor{green}{\ding{51}}& 18.23& 75.67&  57.44&consistent& 96.69& 8.02& 88.67&consistent\\
 Mistral-7B$_{EG}$ \textcolor{red}{\ding{55}}& 18.39& 15.66&  \textbf{-2.73}&\textit{\underline{adversarial}}& 28.49& 29.88& \textbf{-1.39}&\textit{\underline{adversarial}}\\
 \cline{2-9}LLaMA-3-70B$_{EG}$ \textcolor{green}{\ding{51}}& 13.83& 103.23&  89.4&consistent& 71.65& 9.76& 61.89&consistent\\
 LLaMA-3-70B$_{EG}$ \textcolor{red}{\ding{55}}& 23.60& 16.02&  \textbf{-7.58}&\textit{\underline{adversarial}}& 19.67& 64.88& \textbf{-45.21}&\textit{\underline{adversarial}}\\ \hline
\end{tabular}
}
    \caption{Frequency of \textcolor{blue}{premise} and \textcolor{red}{hypothesis} when evaluating the Levy/Holt dataset using different LLMs and their EG-tuned variants ($_{EG}$). \textcolor{green}{\ding{51}} indicates correct predictions, while \textcolor{red}{\ding{55}} indicates incorrect predictions.}
    \label{tab:error_analysis_across_different_LLMs}
    \vspace{-0.1in}
\end{table*}

\subsection{Fine-tune LLMs}
\label{sec:fine-tune llms}
We fine-tune several widely used LLMs on NLI datasets, including DeepSeek-R1-Distill-Llama-8B \cite{deepseekai2025deepseekr1incentivizingreasoningcapability}, Mistral-7B, LLaMA-3-8B-instruct, and LLaMA-3-70B-instruct.
Fine-tuning is conducted using LoRA \cite{hu2022lora} within the PEFT framework \cite{ding2023parameter}, with a learning rate of 1e$^{-4}$, 12 training epochs, rank 8, and a dropout rate of 0.05.  
\begin{table*}[h]
    \centering\resizebox{0.85\textwidth}{!}{
   \begin{tabular}{ccccccc}
\hline
           & \multicolumn{3}{c}{\textbf{Label} = \texttt{Entail}}             & \multicolumn{3}{c}{\textbf{Label} = \texttt{No-Entail}}            \\ \hline
           & premise& hypothesis& \multicolumn{1}{l|}{\textit{Bias(\textcolor{red}{hypo},\textcolor{blue}{prem})}} & premise& hypothesis& \multicolumn{1}{l}{\textit{Bias(\textcolor{blue}{prem},\textcolor{red}{hypo})}} \\
 \cline{2-7} DeepSeek-8B \textcolor{green}{\ding{51}}& 20.27& 65.84& 45.57& 88.98& 11.75&77.23\\
 DeepSeek-8B \textcolor{red}{\ding{55}}& 8.27  & 52.59 & \textbf{44.32}& 40.83 & 24.15 &\textbf{16.68}\\ \hline
 DeepSeek-8B$_{EG}$ \textcolor{green}{\ding{51}}& 15.78& 79.85& 64.07& 95.18& 8.26&86.92\\
DeepSeek-8B$_{EG}$ \textcolor{red}{\ding{55}}& 25.64 & 15.55 & \textbf{-10.09}& 26.50 & 30.06 & \textbf{-3.56}\\ \hline
 DeepSeek-8B$_{RTE}$ \textcolor{green}{\ding{51}}& 21.17& 75.42& 54.25& 73.99& 10.86&63.13\\
 DeepSeek-8B$_{RTE}$ \textcolor{red}{\ding{55}}& 12.97& 42.14& \textbf{29.17}& 44.88& 31.68&\textbf{13.2}\\ \hline
 DeepSeek-8B$_{MNLI}$ \textcolor{green}{\ding{51}}& 19.62& 71.14& 51.52& 86.89& 8.86&78.03\\
 DeepSeek-8B$_{MNLI}$ \textcolor{red}{\ding{55}}& 9.29& 13.88& \textbf{4.59}& 51.33& 23.25&\textbf{28.08}\\ \hline
\end{tabular}
}
    \caption{Frequency of \textcolor{blue}{premise} and \textcolor{red}{hypothesis} when evaluating the Levy/Holt using LLMs fine-tuned on different NLI datasets. \textcolor{green}{\ding{51}} denotes correct predictions, while \textcolor{red}{\ding{55}} indicates incorrect predictions.}
    \label{tab:error_analysis_across_different_train_set}
    \vspace{-0.1in}
\end{table*}

\section{Results}
\label{sec:results}
\subsection{Finding 1: Frequency bias in NLI datasets}
\label{sec:frequency bias exists in Inference}
We analyze average predicate frequency on the various NLI datasets. From Table~\ref{tab:average_frequency_across_P-H}, we observe a consistent pattern: for instances labeled \texttt{Entail}, the hypothesis typically contains predicates with higher corpus frequency. Conversely, for \texttt{No-Entail} instances, predicates in the hypothesis tend to be less frequent. 
This observation present a consistent \textit{\textbf{frequency bias}} embedded in these NLI datasets. The presence of frequency biases in NLI datasets raise the potential possibility that LLMs may be learning and leveraging these frequency patterns during fine-tuning.

\subsection{Finding 2: LLMs are sensitive to frequency bias}
\label{sec:limitaion of LLMs}
We evaluate a range of LLMs and their EGs-tuned variants on Levy/Holt. Model predictions are classified as either \textit{correct} (\textcolor{green}{\ding{51}}) or \textit{incorrect} (\textcolor{red}{\ding{55}}), and we analyze frequency bias across these categories.

We reports the frequency biases in Table~\ref{tab:error_analysis_across_different_LLMs}. The results reveal a clear trend: both standard and fine-tuned LLMs can make \textit{correct} predictions when test samples are consistent with the frequency bias. In contrast, when this bias is reduced or adversarial, the likelihood of \textit{incorrect} predictions increases. It shows that LLMs' performance is sensitive to the frequency bias. Furthermore, we observe that the sensitivity is especially pronounced in EG-tuned models, where the frequency bias is higher in \textit{correct} predictions and more reduced in \textit{incorrect} ones, compared to standard LLMs.

We also evaluate LLMs fine-tuned on different NLI datasets. As shown in Table~\ref{tab:error_analysis_across_different_train_set}, we observe that models consistently tend to make correct predictions on samples with stronger frequency bias, but are more likely to produce incorrect predictions when the bias is reduced. The phenomenon is consistently more pronounced in LLMs fine-tuned on NLI datasets compared to standard models. 

\begin{figure}[!t]
	\centering
	\includegraphics[width=0.9\linewidth]{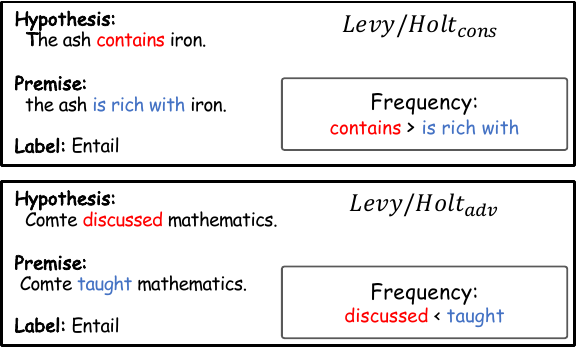}
	\caption{A sample in  Levy/Holt$_{cons}$ and Levy/Holt$_{adv}$.}
	\label{Fig:examples}
 \vspace{-0.1in}
\end{figure}

\subsection{Finding 3: NLI-tuned LLMs struggle with frequency-adversarial inference}
\label{sec:test on adv-Freq Levy/Holt}
To further measure the impact of frequency bias, we divide the Levy/Holt dataset into two subsets: those \textit{consistent} with the frequency bias (Levy/Holt$_{cons}$) and those that are \textit{adversarial} to it (Levy/Holt$_{adv}$). As illustrated in Figure~\ref{Fig:examples}, in Levy/Holt$_{cons}$ samples labeled \texttt{Entail}, the predicate in the hypothesis is more frequent than that in the premise. In contrast, Levy/Holt$_{adv}$ contains cases, where the premise is more frequent than the hypothesis.

\begin{table}[t]
     \centering\resizebox{0.48\textwidth}{!}{
\begin{tabular}{cccc}
\hline
Models          &Levy/Holt$_{cons}$&Levy/Holt$_{adv}$
 &$\Delta$\\ \hline
LLaMA-3-8B      & 74.0 & 61.74 & -12.26\\
DeepSeek-8B  &
73.51&64.99 &-8.52\\
 Mistral-7B& 65.31&57.23&-8.08\\
LLaMA-3-70B     &
84.25&70.55 &-13.7\\ \hline
 LLaMA-3-8B$_{EG}$& 85.2 & 62.5 &\textbf{-22.7}\\
 DeepSeek-8B$_{EG}$& 80.8&62.18 &\textbf{-18.62}\\
 Mistral-7B$_{EG}$& 
83.61&58.79 &\textbf{-24.82}\\
 LLaMA-3-70B$_{EG}$& 85.25& 69.67 &\textbf{-15.58}\\ \hline
    \end{tabular}}
    \caption{AUC scores on the frequency-consistent and frequency-adversarial Levy/Holt.}
    \label{tab:adv_cons_LH}
    \vspace{-0.1in}
\end{table}

We evaluate model performance separately on Levy/Holt$_{cons}$ and Levy/Holt$_{adv}$, and report the Area Under the Curve (AUC) scores in Table~\ref{tab:adv_cons_LH}. Results show a substantial drop in AUC scores on Levy/Holt$_{adv}$ for both standard and fine-tuned LLMs. This performance gaps on  Levy/Holt$_{cons}$ and Levy/Holt$_{adv}$ suggests that LLMs rely on frequency bias as a shortcut, performing well on reasoning from low-frequency to high-frequency statements but struggling when this pattern is reversed. 
Compared to standard LLMs, EG-tuned LLMs exhibit a more substantial performance gaps, indicating that training on EGs reinforces their reliance on the frequency bias in datasets. We further fine-tune LLMs on additional NLI datasets and observe consistent findings, as shown in Appendix \ref{sec:appendix_NLI_freq_adv}.

These results highlight two key findings: (1) LLMs perform well on bias-consistent cases but struggle with adversarial ones, indicating that they exploit frequency bias as a proxy for inference. 
(2) After fine-tuned on NLI datasets, LLMs exhibit increased reliance on frequency bias. This suggests that fine-tuning process encourages learning frequency-based patterns from these datasets, which reinforces inferences from low-frequency to high-frequency statements while diminishing the ability to reason in the opposite direction.

\subsection{Finding 4: Frequency bias is a proxy for gradient of semantic generalization }
\label{sec:frequency_with_generalization}
To further investigate the relationship between predicate frequency and entailment relation, we analyze the frequency of \textbf{hyponym–hypernym pairs}\footnote{Unlike \citet{mckenna2023smoothing}, we focus on \textit{verb} pairs to examine generalization and specificity, as verbs often involve more abstract and dynamic semantic shifts than nouns.} extracted from WordNet \cite{miller-1994-wordnet}. These pairs represent a specific form of upward semantic entailment, namely generalization, where a more specific concept (hyponym) entails its corresponding general one (hypernym). For example, \textit{``whisper''} is a hyponym of \textit{``talk''}, so the statement \textit{``X whisper to Y''} semantically entails \textit{``X talk to Y''}.

\begin{table}[t]
    \centering\resizebox{0.4\textwidth}{!}{
   \begin{tabular}{ccc}
\hline
          & Hyponyms& Hypernyms\\ \hline
 WordNet&  4.13& 12.18\\ \hline
\end{tabular}
}
    \caption{Average frequency of Hyponym-Hypernym in WordNet. The results prove that more general concepts (hypernym) have higher frequency.
    }
    \label{tab:average_frequency_across_Hyponyms-Hypernyms_in_Wordnet}
    \vspace{-0.1in}
\end{table}

Table~\ref{tab:average_frequency_across_Hyponyms-Hypernyms_in_Wordnet} reports the average frequency of hyponyms and hypernyms separately. Hypernyms occur more frequently than their corresponding hyponyms, indicating that more general concept are more frequent in natural language. This finding suggests that frequency bias may serve as a proxy for the generalization gradient, whereby inferences from lower-frequency to higher-frequency predicates reflect a generalization from specific to more abstract concepts.
In NLI datasets, we also observe that when the label is \texttt{Entail}, hypotheses often contain more hypernyms than corresponding premises, as shown in Appendix~\ref{sec:appendix_hypo_hyer_pairs}. This suggests that most samples in NLI datasets require inference from more specific concepts to more general ones.

These findings offer an explanation for why training LLMs on NLI datasets can serve as a proxy for enhancing inferential capability: LLMs learn frequency biases from datasets during training and these biases align with a generalization gradient that supports entailment from specific to more general concepts. 
Although LLMs can learn frequency bias as a proxy to enhance their inferential capability, learning this bias limits model robustness in frequency-adversarial settings, as observed in \S\ref{sec:test on adv-Freq Levy/Holt}.

\section{Conclusion}
\label{sec:conclusion}
In this work, we investigate what LLMs actually learn during the fine-tuning process. First, we identify significant frequency bias in various NLI datasets. Next we prove that LLMs exploit this bias for inference, and that fine-tuning further increases their reliance on such patterns. 
Finally, we show a strong correlation between frequency bias and a particular variety of entailment that is common in NLI datasets, namely hyponym-to-hypernym generalization. It offers an explanation for why learning these frequency patterns can enhance model inference performance. Our work also reveals that a key limitation of LLMs is their vulnerability to frequency-adversarial inference cases. 

\section*{Limitations}
\label{sec:Limitations}
In this work, our findings suggest that LLMs internalize frequency biases from training data and utilize them as a proxy for inference. However, due to computational constraints, our experiments are limited to a range of smaller LLMs with 8B variants and a single extremely large-scale model, LLaMA-3-70B. This restricts our ability to draw conclusions about extremely large models. In future work, we will explore broader model scales to assess the consistency of these trends.

\bibliography{ref}

\begin{thebibliography}{23}
\expandafter\ifx\csname natexlab\endcsname\relax\def\natexlab#1{#1}\fi

\bibitem[{Berant et~al.(2010)Berant, Dagan, and Goldberger}]{berant2010global}
Jonathan Berant, Ido Dagan, and Jacob Goldberger. 2010.
\newblock {Global Learning} of {Focused Entailment Graphs}.
\newblock In \emph{Proceedings of the 48th Annual Meeting of the Association for Computational Linguistics}, pages 1220--1229.

\bibitem[{Berant et~al.(2011)Berant, Dagan, and Goldberger}]{berant2011global}
Jonathan Berant, Ido Dagan, and Jacob Goldberger. 2011.
\newblock {Global} {Learning} of {Typed} {Entailment} {Rules}.
\newblock In \emph{Proceedings of the 49th Annual Meeting of the Association for Computational Linguistics: Human Language Technologies}, pages 610--619.

\bibitem[{Cheng et~al.(2025)Cheng, Li, Wang, Liu, and Steedman}]{cheng2025neutralizing}
Liang Cheng, Tianyi Li, Zhaowei Wang, Tianyang Liu, and Mark Steedman. 2025.
\newblock Neutralizing bias in llm reasoning using entailment graphs.
\newblock \emph{arXiv preprint arXiv:2503.11614}.

\bibitem[{Dagan et~al.(2006)Dagan, Glickman, and Magnini}]{dagan2006pascal}
Ido Dagan, Oren Glickman, and Bernardo Magnini. 2006.
\newblock The {PASCAL} recognising textual entailment challenge.
\newblock In \emph{Machine learning challenges. evaluating predictive uncertainty, visual object classification, and recognising tectual entailment}, pages 177--190. Springer.

\bibitem[{DeepSeek-AI et~al.(2025)DeepSeek-AI, Guo, Yang, Zhang, Song, Zhang, Xu, Zhu, Ma, Wang, Bi, Zhang, Yu, Wu, Wu, Gou, Shao, Li, Gao, Liu, Xue, Wang, Wu, Feng, Lu, Zhao, Deng, Zhang, Ruan, Dai, Chen, Ji, Li, Lin, Dai, Luo, Hao, Chen, Li, Zhang, Bao, Xu, Wang, Ding, Xin, Gao, Qu, Li, Guo, Li, Wang, Chen, Yuan, Qiu, Li, Cai, Ni, Liang, Chen, Dong, Hu, Gao, Guan, Huang, Yu, Wang, Zhang, Zhao, Wang, Zhang, Xu, Xia, Zhang, Zhang, Tang, Li, Wang, Li, Tian, Huang, Zhang, Wang, Chen, Du, Ge, Zhang, Pan, Wang, Chen, Jin, Chen, Lu, Zhou, Chen, Ye, Wang, Yu, Zhou, Pan, Li, Zhou, Wu, Ye, Yun, Pei, Sun, Wang, Zeng, Zhao, Liu, Liang, Gao, Yu, Zhang, Xiao, An, Liu, Wang, Chen, Nie, Cheng, Liu, Xie, Liu, Yang, Li, Su, Lin, Li, Jin, Shen, Chen, Sun, Wang, Song, Zhou, Wang, Shan, Li, Wang, Wei, Zhang, Xu, Li, Zhao, Sun, Wang, Yu, Zhang, Shi, Xiong, He, Piao, Wang, Tan, Ma, Liu, Guo, Ou, Wang, Gong, Zou, He, Xiong, Luo, You, Liu, Zhou, Zhu, Xu, Huang, Li, Zheng, Zhu, Ma, Tang, Zha, Yan, Ren, Ren, Sha, Fu, Xu, Xie, Zhang,
  Hao, Ma, Yan, Wu, Gu, Zhu, Liu, Li, Xie, Song, Pan, Huang, Xu, Zhang, and Zhang}]{deepseekai2025deepseekr1incentivizingreasoningcapability}
DeepSeek-AI, Daya Guo, Dejian Yang, Haowei Zhang, Junxiao Song, Ruoyu Zhang, Runxin Xu, Qihao Zhu, Shirong Ma, Peiyi Wang, Xiao Bi, Xiaokang Zhang, Xingkai Yu, Yu~Wu, Z.~F. Wu, Zhibin Gou, Zhihong Shao, Zhuoshu Li, Ziyi Gao, Aixin Liu, Bing Xue, Bingxuan Wang, Bochao Wu, Bei Feng, Chengda Lu, Chenggang Zhao, Chengqi Deng, Chenyu Zhang, Chong Ruan, Damai Dai, Deli Chen, Dongjie Ji, Erhang Li, Fangyun Lin, Fucong Dai, Fuli Luo, Guangbo Hao, Guanting Chen, Guowei Li, H.~Zhang, Han Bao, Hanwei Xu, Haocheng Wang, Honghui Ding, Huajian Xin, Huazuo Gao, Hui Qu, Hui Li, Jianzhong Guo, Jiashi Li, Jiawei Wang, Jingchang Chen, Jingyang Yuan, Junjie Qiu, Junlong Li, J.~L. Cai, Jiaqi Ni, Jian Liang, Jin Chen, Kai Dong, Kai Hu, Kaige Gao, Kang Guan, Kexin Huang, Kuai Yu, Lean Wang, Lecong Zhang, Liang Zhao, Litong Wang, Liyue Zhang, Lei Xu, Leyi Xia, Mingchuan Zhang, Minghua Zhang, Minghui Tang, Meng Li, Miaojun Wang, Mingming Li, Ning Tian, Panpan Huang, Peng Zhang, Qiancheng Wang, Qinyu Chen, Qiushi Du, Ruiqi Ge, Ruisong
  Zhang, Ruizhe Pan, Runji Wang, R.~J. Chen, R.~L. Jin, Ruyi Chen, Shanghao Lu, Shangyan Zhou, Shanhuang Chen, Shengfeng Ye, Shiyu Wang, Shuiping Yu, Shunfeng Zhou, Shuting Pan, S.~S. Li, Shuang Zhou, Shaoqing Wu, Shengfeng Ye, Tao Yun, Tian Pei, Tianyu Sun, T.~Wang, Wangding Zeng, Wanjia Zhao, Wen Liu, Wenfeng Liang, Wenjun Gao, Wenqin Yu, Wentao Zhang, W.~L. Xiao, Wei An, Xiaodong Liu, Xiaohan Wang, Xiaokang Chen, Xiaotao Nie, Xin Cheng, Xin Liu, Xin Xie, Xingchao Liu, Xinyu Yang, Xinyuan Li, Xuecheng Su, Xuheng Lin, X.~Q. Li, Xiangyue Jin, Xiaojin Shen, Xiaosha Chen, Xiaowen Sun, Xiaoxiang Wang, Xinnan Song, Xinyi Zhou, Xianzu Wang, Xinxia Shan, Y.~K. Li, Y.~Q. Wang, Y.~X. Wei, Yang Zhang, Yanhong Xu, Yao Li, Yao Zhao, Yaofeng Sun, Yaohui Wang, Yi~Yu, Yichao Zhang, Yifan Shi, Yiliang Xiong, Ying He, Yishi Piao, Yisong Wang, Yixuan Tan, Yiyang Ma, Yiyuan Liu, Yongqiang Guo, Yuan Ou, Yuduan Wang, Yue Gong, Yuheng Zou, Yujia He, Yunfan Xiong, Yuxiang Luo, Yuxiang You, Yuxuan Liu, Yuyang Zhou, Y.~X. Zhu,
  Yanhong Xu, Yanping Huang, Yaohui Li, Yi~Zheng, Yuchen Zhu, Yunxian Ma, Ying Tang, Yukun Zha, Yuting Yan, Z.~Z. Ren, Zehui Ren, Zhangli Sha, Zhe Fu, Zhean Xu, Zhenda Xie, Zhengyan Zhang, Zhewen Hao, Zhicheng Ma, Zhigang Yan, Zhiyu Wu, Zihui Gu, Zijia Zhu, Zijun Liu, Zilin Li, Ziwei Xie, Ziyang Song, Zizheng Pan, Zhen Huang, Zhipeng Xu, Zhongyu Zhang, and Zhen Zhang. 2025.
\newblock \href {http://arxiv.org/abs/2501.12948} {Deepseek-r1: Incentivizing reasoning capability in llms via reinforcement learning}.

\bibitem[{Ding et~al.(2023)Ding, Qin, Yang, Wei, Yang, Su, Hu, Chen, Chan, Chen et~al.}]{ding2023parameter}
Ning Ding, Yujia Qin, Guang Yang, Fuchao Wei, Zonghan Yang, Yusheng Su, Shengding Hu, Yulin Chen, Chi-Min Chan, Weize Chen, et~al. 2023.
\newblock Parameter-efficient fine-tuning of large-scale pre-trained language models.
\newblock \emph{Nature Machine Intelligence}, 5(3):220--235.

\bibitem[{He et~al.(2024)He, Wu, Camburu, Minervini, and Stenetorp}]{he2024using}
Xuanli He, Yuxiang Wu, Oana-Maria Camburu, Pasquale Minervini, and Pontus Stenetorp. 2024.
\newblock Using natural language explanations to improve robustness of in-context learning.
\newblock In \emph{Proceedings of the 62nd Annual Meeting of the Association for Computational Linguistics (Volume 1: Long Papers)}, pages 13477--13499.

\bibitem[{Holt(2019)}]{holt_probabilistic_2019}
Xavier Holt. 2019.
\newblock \href {http://arxiv.org/abs/1907.12048} {Probabilistic {Models} of {Relational} {Implication}}.
\newblock \emph{arXiv:1907.12048 [cs, stat]}.
\newblock ArXiv: 1907.12048.

\bibitem[{Hosseini et~al.(2018)Hosseini, Chambers, Reddy, Holt, Cohen, Johnson, and Steedman}]{hosseini2018learning}
Mohammad~Javad Hosseini, Nathanael Chambers, Siva Reddy, Xavier~R Holt, Shay~B Cohen, Mark Johnson, and Mark Steedman. 2018.
\newblock {Learning Typed Entailment Graphs} with {Global Soft Constraints}.
\newblock \emph{Transactions of the Association for Computational Linguistics}, 6:703--717.

\bibitem[{Hosseini et~al.(2021)Hosseini, Cohen, Johnson, and Steedman}]{hosseini2021open}
Mohammad~Javad Hosseini, Shay~B Cohen, Mark Johnson, and Mark Steedman. 2021.
\newblock {Open}-{Domain Contextual Link Prediction} and its {Complementarity} with {Entailment Graphs}.
\newblock In \emph{Findings of the Association for Computational Linguistics: EMNLP 2021}, pages 2790--2802.

\bibitem[{Hu et~al.(2022)Hu, Shen, Wallis, Allen-Zhu, Li, Wang, Wang, and Chen}]{hu2022lora}
Edward~J Hu, Yelong Shen, Phillip Wallis, Zeyuan Allen-Zhu, Yuanzhi Li, Shean Wang, Lu~Wang, and Weizhu Chen. 2022.
\newblock \href {https://openreview.net/forum?id=nZeVKeeFYf9} {Lo{RA}: Low-rank adaptation of large language models}.
\newblock In \emph{International Conference on Learning Representations}.

\bibitem[{Levy and Dagan(2016)}]{levy_annotating_2016}
Omer Levy and Ido Dagan. 2016.
\newblock \href {https://doi.org/10.18653/v1/P16-2041} {Annotating {Relation} {Inference} in {Context} via {Question} {Answering}}.
\newblock In \emph{Proceedings of the 54th {Annual} {Meeting} of the {Association} for {Computational} {Linguistics} ({Volume} 2: {Short} {Papers})}, pages 249--255, Berlin, Germany. Association for Computational Linguistics.

\bibitem[{Li et~al.(2025)Li, Wang, Zhang, Yin, Duan, Xiao, and Tang}]{li2025dynmole}
Dengchun Li, Naizheng Wang, Zihao Zhang, Haoyang Yin, Lei Duan, Meng Xiao, and Mingjie Tang. 2025.
\newblock Dynmole: Boosting mixture of lora experts fine-tuning with a hybrid routing mechanism.
\newblock \emph{arXiv preprint arXiv:2504.00661}.

\bibitem[{Li et~al.(2022)Li, Hosseini, Weber, and Steedman}]{li2022language}
Tianyi Li, Mohammad~Javad Hosseini, Sabine Weber, and Mark Steedman. 2022.
\newblock Language models are poor learners of directional inference.
\newblock In \emph{Findings of the Association for Computational Linguistics: EMNLP 2022}, pages 903--921.

\bibitem[{Liu et~al.(2020)Liu, Cheng, He, Chen, Wang, Poon, and Gao}]{liu2020adversarial}
Xiaodong Liu, Hao Cheng, Pengcheng He, Weizhu Chen, Yu~Wang, Hoifung Poon, and Jianfeng Gao. 2020.
\newblock Adversarial training for large neural language models.
\newblock \emph{arXiv preprint arXiv:2004.08994}.

\bibitem[{Liu et~al.(2024)Liu, Xu, Wu, Yuan, Yang, Zhou, Liu, Guan, Wang, Yu et~al.}]{liu2024large}
Xiaoyu Liu, Paiheng Xu, Junda Wu, Jiaxin Yuan, Yifan Yang, Yuhang Zhou, Fuxiao Liu, Tianrui Guan, Haoliang Wang, Tong Yu, et~al. 2024.
\newblock Large language models and causal inference in collaboration: A comprehensive survey.
\newblock \emph{arXiv preprint arXiv:2403.09606}.

\bibitem[{Mckenna et~al.(2023{\natexlab{a}})Mckenna, Li, Cheng, Hosseini, Johnson, and Steedman}]{mckenna2023sources}
Nick Mckenna, Tianyi Li, Liang Cheng, Mohammad Hosseini, Mark Johnson, and Mark Steedman. 2023{\natexlab{a}}.
\newblock Sources of hallucination by large language models on inference tasks.
\newblock In \emph{Findings of the Association for Computational Linguistics: EMNLP 2023}, pages 2758--2774.

\bibitem[{Mckenna et~al.(2023{\natexlab{b}})Mckenna, Li, Johnson, and Steedman}]{mckenna2023smoothing}
Nick Mckenna, Tianyi Li, Mark Johnson, and Mark Steedman. 2023{\natexlab{b}}.
\newblock Smoothing entailment graphs with language models.
\newblock In \emph{Proceedings of the 13th International Joint Conference on Natural Language Processing and the 3rd Conference of the Asia-Pacific Chapter of the Association for Computational Linguistics (Volume 1: Long Papers)}, pages 551--563.

\bibitem[{Miller(1994)}]{miller-1994-wordnet}
George~A. Miller. 1994.
\newblock \href {https://aclanthology.org/H94-1111/} {{W}ord{N}et: A lexical database for {E}nglish}.
\newblock In \emph{{H}uman {L}anguage {T}echnology: Proceedings of a Workshop held at {P}lainsboro, {N}ew {J}ersey, {M}arch 8-11, 1994}.

\bibitem[{Schmitt and Schütze(2021)}]{schmitt_language_2021}
Martin Schmitt and Hinrich Schütze. 2021.
\newblock \href {https://www.aclweb.org/anthology/2021.eacl-main.108} {Language {Models} for {Lexical} {Inference} in {Context}}.
\newblock In \emph{Proceedings of the 16th {Conference} of the {European} {Chapter} of the {Association} for {Computational} {Linguistics}: {Main} {Volume}}, pages 1267--1280, Online. Association for Computational Linguistics.

\bibitem[{Speer(2022)}]{robyn_speer_2022_7199437}
Robyn Speer. 2022.
\newblock \href {https://doi.org/10.5281/zenodo.7199437} {rspeer/wordfreq: v3.0}.

\bibitem[{Wang et~al.(2019)Wang, Singh, Michael, Hill, Levy, and Bowman}]{wang2019glue}
Alex Wang, Amanpreet Singh, Julian Michael, Felix Hill, Omer Levy, and Samuel~R. Bowman. 2019.
\newblock {GLUE}: A multi-task benchmark and analysis platform for natural language understanding.
\newblock In the Proceedings of ICLR.

\bibitem[{Williams et~al.(2018)Williams, Nangia, and Bowman}]{williams2018broad}
Adina Williams, Nikita Nangia, and Samuel~R. Bowman. 2018.
\newblock A broad-coverage challenge corpus for sentence understanding through inference.
\newblock In \emph{Proceedings of NAACL-HLT}.

\end{thebibliography}
\bibliographystyle{acl_natbib}

\newpage
\appendix
\section{Dataset}
\label{sec:appendix_dataset}

\subsection{Train set}
\label{sec:appendix_train_set}
\paragraph{Entailment Graphs (EGs)} \cite{berant2010global,berant2011global, hosseini2018learning, hosseini2021open} are symbolic graphs to preserve the textual entailment in open-domain corpora. These graphs contain entailment between predicates, formatted as triples consisting of predicate pairs and their typed arguments.
\citet{cheng2025neutralizing} propose a method to initialize the extracted EGs into inference datasets for training LLMs. The datasets comprise premise–hypothesis pairs, with each hypothesis being counterfactual yet logically entailed by its corresponding premise. Fine-tuning LLMs on this data has been shown to substantially enhance inferential capability while reducing hallucinations \cite{cheng2025neutralizing}. 

\paragraph{RTE} \cite{dagan2006pascal, wang2019glue} is a NLI benchmark dataset designed for evaluating models on the task of recognizing textual entailment (RTE). It consists of sentence-level premise-hypothesis pairs collected from various sources, which are collected from various sources such as news articles and information extraction tasks.

\paragraph{MNLI} \cite{williams2018broad, wang2019glue} is a widely used benchmark for evaluating language models on NLI, It consists of sentence-level premise-hypothesis pairs, with premises drawn from diverse sources and hypotheses manually written.
In our experiments, we fine-tune LLMs using the MNLI training split.

\subsection{Test set}
\label{sec:appendix_test set}
\paragraph{Levy/Holt} \cite{levy_annotating_2016, holt_probabilistic_2019} dataset is a widely used for NLI, which comprises premise-hypothesis pairs structured in a specific task format: ``Given [premise $P$], is it true that [hypothesis $H$]?''. Each $P$- and $H$-statement has the property of containing one predicate with two named entity arguments, where the same entities appear in both $P$ and $H$. The Levy/Holt dataset contains inverse of all entailment pairs. In our experiments, we study the challenging directional subset, where the entailments hold in one direction but not both.

\section{Prompts Used in Experiments}
\label{sec:appendix_prompts}
\paragraph{Prompt templates} are widely acknowledged for their significant and sometimes decisive impact on the behavior of LLMs. In our experiments, we categorize the prompt templates into two distinct types based on their usage: prompt templates for fine-tuning LLMs and prompt templates used during inference. 

\subsection{prompt template for fine-tuning}
\label{sec:appendix_prompt template for fine-tuning}
We follow the fine-tuning setup of \citet{cheng2025neutralizing}, adopting the same prompt templates used in prior inference studies \cite{schmitt_language_2021, mckenna2023sources, cheng2025neutralizing}, which follow the format outlined below: 

\begin{enumerate}
    \item[] If $[$\textsc{premise}$]$, then $[$\textsc{hypothesis}$]$.
\end{enumerate}

 To make LLMs better understanding the task, we format it as Boolean questions and include indicator words such as ``Question:" and ``Answer:". For each option, we automatically provide explanations for every answer by adding affirmation or negation to the propositions. As a result, the NLI training data is structured as shown in Table~\ref{Tab:prompt_for_training}. We fine-tune our models using these templates. 

\subsection{prompt template for inference}
\label{sec:appendix_prompt template for attesting inference}
Following the evaluation settings of prior works \cite{schmitt_language_2021, mckenna2023sources, cheng2025neutralizing}, we use the same few-shot examples in our inference prompts for NLI tasks, consisting of two positive and two negative instances. The examples are shown in Table~\ref{tab:example_instantiated_prompts}.

\section{Hyponymn-Hypernym Pairs in NLI datasets}
\label{sec:appendix_hypo_hyer_pairs}
We analyze the distribution of hyponym–hypernym pairs in the NLI dataset by counting the occurrences of hypernyms and hyponyms in premises and hypotheses. We focus our analysis on the EG and Levy/Holt datasets because their samples contain explicit predicate structures, unlike MNLI and RTE, which are at the sentence level and lack clearly defined predicates. 
As shown in Table~\ref{tab:Hyponyms-Hypernyms_in_LevyHolt}, we observe that for instances labeled as \texttt{Entail}, hypernyms appear more frequently in the hypothesis than in the premise. Conversely, for \texttt{No-Entail} instances, hypernyms are more frequent in the premise. These results suggest that hypotheses tend to contain more abstract concepts than premises in positive examples, aligning with the observed frequency bias and reinforcing the findings discussed in \S\ref{sec:frequency_with_generalization}. 

\begin{table}[t]
\centering\resizebox{0.49\textwidth}{!}{\begin{tabular}{ccccc}
\hline
          & \multicolumn{2}{c|}{when Label = Entail}  & \multicolumn{2}{c}{when Label = No-Entail} \\ \hline
          & Hypernmys & \multicolumn{1}{c|}{Hyponyms} & Hypernmys            & Hyponyms            \\ \cline{2-5} 
Levy/Holt & 92& 65& 56& 101\\
 EG& 93& 64& 31&61\\ \hline
\end{tabular}}
 \caption{Counts of Hypernyms and Hyponyms in hypotheses and premises of Levy/Holt and EGs.
We extract all hyponym–hypernym pairs from the data. For samples labeled as \texttt{Entail}, hypotheses tend to contain more hypernyms, indicating a more general statement. In contrast, for \texttt{No-Entail} samples, premises typically include more hypernyms, suggesting that the corresponding hypotheses are more specific or less general.}
    \label{tab:Hyponyms-Hypernyms_in_LevyHolt}
    \vspace{-0.1in}
\end{table}

\section{Fine-tuned LLMs Performance on Frequency-adversarial Inference }
\label{sec:appendix_NLI_freq_adv}
Table~\ref{tab:adv_score_across_diiferent_train_model} presents the performance of LLMs fine-tuned on various NLI datasets, consistently showing that these models struggle with frequency-adversarial inference. 
\begin{table}[t]
    \centering\resizebox{0.5\textwidth}{!}{
\begin{tabular}{cccc}
\hline
Models          &Levy/Holt$_{cons}$&Levy/Holt$_{adv}$
 &$\Delta$\\ \hline
DeepSeek-R1-8B  &
73.51&64.99 &-8.52\\
 DeepSeek-R1-8B$_{EG}$& 80.8&62.18 &\textbf{-18.62}\\
 DeepSeek-R1-8B$_{RTE}$& 73.12& 61.74&-11.38\\
 DeepSeek-R1-8B$_{MNLI}$& 74.09& 61.85&-12.24\\ \hline
 LLaMA-3-8B& 74.0& 61.74&-12.26\\
 LLaMA-3-8B$_{EG}$& 85.2& 62.5&\textbf{-22.7}\\
 LLaMA-3-8B$_{RTE}$& 73.97 & 57.49& -16.48\\
 LLaMA-3-8B$_{MNLI}$& 72.33 & 58.91 & -13.42\\ \hline
    \end{tabular}}
    \caption{AUC scores on the frequency-consistent and frequency-adversarial Levy/Holt.}
    \label{tab:adv_score_across_diiferent_train_model}
\end{table}

\section{Computing Costs}
We fine-tuned the LLaMA-3-70B model on the EGs dataset using four NVIDIA RTX A6000 GPUs over 21 hours. Fine-tuning on the MNLI dataset required approximately 46 hours. For inference, evaluation on the Levy/Hot datasets takes around 30 minutes.

 \begin{table*}[t]
        \normalsize
	   \centering
\begin{tabular}{ll}
\hline
              & \begin{tabular}[c]{@{}l@{}}Question: If $[$\textsc{premise}$]$, then $[$\textsc{hypothesis}$]$. Is that true or false?\\ (A) True; (B) false\end{tabular} \\ \hline
label=\textit{True}  & \begin{tabular}[c]{@{}l@{}}(A) True.\\ Yes, it is true. $[$\textsc{premise}$]$ entails $[$\textsc{hypothesis}$]$.\end{tabular}                            \\ \hline
label=\textit{False} & \begin{tabular}[c]{@{}l@{}}(B) False.\\ No, it is false. $[$\textsc{premise}$]$ does not entail $[$\textsc{hypothesis}$]$.\end{tabular}                   \\ \hline
\end{tabular}
\caption{The table present the prompt template using in our training steps.}
	\label{Tab:prompt_for_training}
\end{table*}

\begin{table*}[t]
    \centering
    \begin{tabular}{p{12cm}}
        \toprule \hline
        \textbf{Few-shot Examples Instantiated Prompt for Inference Task}\\ \hline
        \midrule
        If Google bought Youtube, then Google owns Youtube. Is that true or false?\\
A) True\\
B) False\\
Answer: A) True. Owning is a consequence of buying.\\
If Google owns Youtube, then Google bought Youtube. Is that true or false?\\
A) True\\
B) False\\
Answer: B) False. Owning does not imply buying, the ownership may come from other means.\\
If John went to the mall, then John drove to the mall. Is that true or false?\\
A) True\\
B) False\\
Answer: B) False. John may have gone to the mall by other means.\\
If John drove to the mall, then John went to the mall. Is that true or false?\\
A) True\\
B) False\\
Answer: A) true. Driving is a means of going to the mall.\\
If John F. Kennedy was killed in Dallas, then John F. Kennedy died in Dallas. Is that true or false?\\
A) True\\
B) False\\
Answer:\\\hline
    \end{tabular}
    \caption{Example instantiated prompts in Few-shot settings, for the sample ``\textsc{premise}: [Google bought Youtube], \textsc{hypothesis}: [Google owns Youtube]''.}
    \label{tab:example_instantiated_prompts}
\end{table*}

\end{document}